\documentclass[a4paper,11pt]{article}

\pdfoutput=1
\usepackage{booktabs}

\usepackage{lineno}
\usepackage[colorlinks=false]{hyperref}
\usepackage{bm}
\usepackage{graphicx}
\usepackage{amssymb}
\usepackage{amsmath}
\usepackage{tabularx}
\usepackage{natbib}
\usepackage{floatrow}
\usepackage{caption}
\usepackage{amsmath}
\usepackage{amsfonts}
\usepackage{amssymb}

\usepackage{xcolor}
\usepackage{subfig}
\usepackage{graphicx,amsmath,bm}
\usepackage[binary-units=true]{siunitx}
\usepackage{booktabs}
\usepackage{array}
\usepackage{authblk}
\usepackage{rotating}
\usepackage[left=15mm,right=15mm,top=1.5cm,bottom=1.5cm,includeheadfoot]{geometry}
\setcitestyle{square,numbers}
\begin{document}

\title{Physics-guided Loss Functions Improve Deep Learning Performance in Inverse Scattering}

\author[1,2]{Zicheng~Liu}
\author[3]{Mayank~Roy}
\author[3]{Dilip~K.~Prasad}
\author[2]{Krishna~Agarwal}

\affil[1]{\scriptsize Department of Electronic Engineering, School of Electronics and Information, Northwestern Polytechnical University, Xi'an 710029, China}
\affil[2]{\scriptsize Department of Physics and Technology, UiT The Arctic University of Norway, NO-9037 Troms\o, Norway}
\affil[3]{\scriptsize Department of Computer Science, UiT The Arctic University of Norway, NO-9037 Troms\o, Norway}

\maketitle

\abstract{
Solving electromagnetic inverse scattering problems (ISPs) is challenging due to the intrinsic nonlinearity, ill-posedness, and expensive computational cost. Recently, deep neural network (DNN) techniques have been successfully applied on ISPs and shown potential of superior imaging over conventional methods. In this paper, we analyse the analogy between DNN solvers and traditional iterative algorithms and discuss how important physical phenomena cannot be effectively incorporated in the training process. We show the importance of including near-field priors in the learning process of DNNs. To this end, we propose new designs of loss functions which incorporate multiple-scattering based near-field quantities (such as scattered fields or induced currents within domain of interest). Effects of physics-guided loss functions are studied using a variety of numerical experiments. Pros and cons of the investigated ISP solvers with different loss functions are summarized.}

\section{Introduction}
%
%
%
%
Due to the ability of noninvasive imaging of internal structures, electromagnetic inverse scattering problems (ISPs)\cite{chen2018computational} are widely studied in industrial applications including remote sensing, biomedical imaging, non-destructive testing and geophysics. Images are obtained by collecting the fields scattered by the objects and reconstructing the distribution of electromagnetic constitutive parameters (relative permittivity and conductivity) of the objects. The phenomenon of multiple scattering leads to nonlinear reconstruction problems, which present additional complexity over the inherent ill-posedness of the inverse problem. 

The conventional solvers for non-linear inverse scattering problems include deterministic methods (including distorted Born iterative method \cite{chew1990reconstruction,jun1999convergence}, contrast source inversion \cite{van1997contrast}, and subspace-based optimization method \cite{chen2009subspace,agarwal2010subspace}) and stochastic methods such as genetic optimization \cite{pastorino2000microwave} and differential evolution \cite{agarwal2007application}. Moreover, diverse regularization approaches and prior information have been widely applied to overcome the ill-posedness of ISPs \cite{oliveri2017compressive,shen2014sar,liu2018electromagnetic,anselmi2018iterative}. Since iterative schemes are usually followed and a forward modeling problem needs to be solved at each iteration, the time complexity of such approaches is prohibitive for real-time applications. For the ease of reference, we refer to this family of conventional iterative solvers for ISPs as `traditional ISP solvers'. 

Deep neural network (DNN) has been successfully applied in image processing, computer vision, nature language processing, and electromagnetic computation. Recently, its application on ISPs \cite{chen2020review,xu2020deep,ran2020subwavelength,rekanos2002neural} has drawn  much attention due to the potential of superior performances in imaging accuracy and efficiency. In the learning-by-examples paradigm \cite{ahmed2019adaptive,salucci2019nonlinear}, electric fields are taken as input and the technique of principal component analysis can be used to reduce the input dimension and avoid the problem of curse-of-dimensionality. Convolutional neural networks (CNNs) \cite{ran2020imaging} with U-net architecture \cite{ronneberger2015u,jin2017deep,wei2018deep} have also been used. 

\begin{figure}[!ht]
	\centering
	\includegraphics[width = 0.7\linewidth]{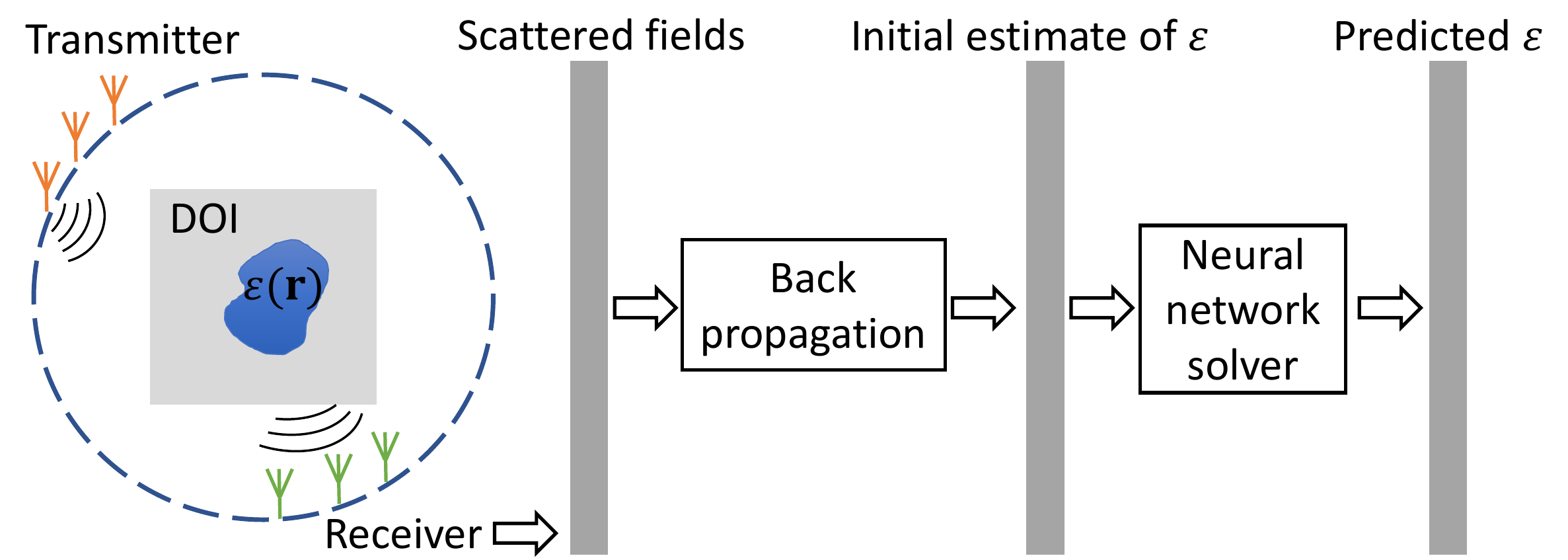}
	\caption{Sketch of the imaging system and deep neural network (DNN) solver for the concerned inverse scattering problems. Fields scattered by objects in domain of interest (DOI) are collected by receivers and used to have an initial estimate of contrast function through back propagation method. The DNN solver takes the initial estimate as input and generates the final estimate of contrast function.}
	\label{fig:sketchISP}
\end{figure}

Some incorporation of physics-guided priors have been explored within the framework of DNN. Taking estimations generated by fast non-iterative conventional methods (e.g., back propagation) as input and the ground truth as the response, the predictor is trained based on simulation results and then used for real scenarios \cite{wei2018deep,li2018deepnis,xu2020deep}. See Fig. \ref{fig:sketchISP}. In contrast with directly predicting the contrast (or permittivity) function for most DNN solvers, an induced current learning method (ICLM) \cite{wei2019physics} has also been proposed. Based on the estimated induced current density, the contrast function is computed following a deterministic approach. In \cite{yao2019two}, the DNN solver is designed to solve ISPs in two steps: an initial estimate of contrast is generated by a CNN which takes scattered fields as input and then the final solution is obtained further learning the solution residue using a modified U-Net. For the above DNN solvers, it is not clear how physical information is considered in the training process. Therefore, it is hard to gain insight or perform educated manipulation of such approaches for generalizations. 

An attempt to draw an analogy between DNN solvers and the conventional iterative algorithms has been made recently \cite{li2018deepnis}. This analogy is helpful to us for providing solutions that integrate the physics of the ISPs with DNN through design of physics-guided loss functions. In general, ISPs can be transformed as regularized optimization problems which can be iteratively solved. While the recursive formula applied in iterative algorithms may be equivalent to a full-connected layer of DNN, the iteration starts from the initial solution which is the input contrast function of the DNN. Following this analogy and analyzing DNN solvers from the view of iterative algorithms, the loss function that penalizes discrepancies in dielectric contrast distributions (or analogously scattering contrasts) guides the network to have predictions that fit the patterns of the scattering contrast distributions in the training set. In other words, the network is actually inclined to learn only how to regularize the desired scattering contrast function. 

An underlying insight here is that the scattering contrast is intrinsic property of the scatterers for a given background and it therefore does not incorporate the extrinsic manifestation of the physics of scattering which ultimately results into the measured scattered field. In this sense, it is useful to consider if loss function can be designed to utilize the external manifestations of the physical phenomenon of scattering and thereby fit more than just the scattering contrast distributions. 

Noise adversely affects the prediction accuracy and the robustness of DNN solvers. In the training process, the network parameters (weights and biases) are optimized by minimizing the loss function, which usually measures the discrepancies between the response and the ground truth. If clean input (i.e., initial estimation from noise-free scattered fields) is used for training, the training process has no chance to consider the effects from noises. Therefore, the obtained solver may behave poorly in real scenarios. While it is possible to consider the noise effects in the learning process by generating input images from noisy fields, further improvements can be achieved by incorporating the noise effects in the loss function. In the traditional ISP solvers \cite{oliveri2017compressive,liu2018electromagnetic}, effect of noise is considered by including the so-called data-fitting term, which is missing in the DNN loss function. This data fitting term works against noise by assessing if the estimated scatterers match the electric far-field distribution as well. While such data-fitting term may be included in DNN loss-function as well, we target near-field physics of scattering for designing the loss functions. 

The source of our inspiration lies in the fact that near-field multiple scattering phenomenon is the root cause of non-linearity of ISPs as well as a significant contributor of ill-posedness because it cannot be tapped in the far-field measurements. This aspect has been powerfully used in a special category of algorithms within the iterative solvers. While almost all ISP algorithms within this family use data-fitting loss function, the algorithms referred to here incorporate a loss term related to the constraint on the near-field quantities within the domain of interest (DOI), usually through some component of the near-field state equation related to multiple scattering \cite{chen2018computational}. The relevant near field quantities may be scattered field , induced currents or total electric field within the DOI, and it has been consistently demonstrated that the inclusion of near-field loss function terms are critical for improvement in the accuracy of reconstruction \cite{yin2018wavelet}. While Wei et. al \cite{wei2019physics} attempted to incorporate such information in their DNN approach by first reconstructing the induced currents in the DOI using a DNN solver, incorporating such information through loss functions of DNN and thereby introducing the helpful near-field priors in learning is still pending.

In this paper, two variations of physics-guided loss functions are proposed. They incorporate multiple scattering through near-field quantities, namely induced current distribution and scattered electric field distribution in the DOI. The induced current distribution is zero in the background, and this information incorporates the geometry of the scatterers. The actual induced current distribution on the scatterer further encodes multiple scattering if computed using non-linear multiple scattering model, thereby incorporating physical constraints on the dielectric contrast and refining the solution space to incorporate only those physically-viable solutions that match both the geometry and the multiple-scattering imposed current distributions. This corresponds to feature-enhanced learning in the parlance of deep learning. The design of loss function using scattered near-field is proposed for the situations of poor signal-to-noise ratios (SNR). In such situations, small artifacts are likely to appear in the reconstructed scatter profiles such that the induced currents from these artifacts compensate for the noise in the measurements. Here, the scattered near-field provide a better constraint since the artifacts would contribute scattered electric field hotspots that contradict with the expected scattered field profile computed using non-linear multiple scattering model. Therefore, a loss function that utilizes scattered near-field would penalize such solutions. Such a loss function corresponds to improved robustness of the performance of DNN.

This paper is organized as follows. Section \ref{sec:formulateISP} introduces the general framework employed by the traditional ISPs. In Section \ref{sec:DNN}, we discuss the analogy between neural network solvers and iterative schemes and present the two modified loss functions for use in DNN. Numerical results are given in Section \ref{sec:results} and conclusions are made in Section \ref{sec:Conclusions}. Notations for vectors or matrices are in bold, while scalar quantities are in italics. Other sub- and superscripts that abbreviate descriptions are in normal font.

\section{Formulation of the inverse scattering problem}
\label{sec:formulateISP}

Two-dimensional scattering problem is considered with the imaging system sketched in Fig.~\ref{fig:sketchISP}. Transverse-magnetic (TM) waves illuminate scatterers whose longitudinal direction is perpendicular to the plane of paper. Thus, the scope of the current paper is the scalar electromagnetic scattering problem. The scattering behavior can be modeled using the Lippmann-Schwinger equation \cite{chen2018computational}:

\begin{equation}
	\mathbf{E}^\text{tot}_\text{DOI}(\mathbf{r}) = \mathbf{E}^\text{inc}_\text{DOI}(\mathbf{r}) + k_0^2\int_\text{DOI}g(\mathbf{r},\mathbf{r}^\prime)\mathbf{J}(\mathbf{r}^\prime)d\mathbf{r}^\prime, \,\, \mathbf{r}\in\text{DOI},
	\label{eq:stateEqu}
\end{equation}
where $\mathbf{E}^\text{tot}_\text{DOI}$ and $\mathbf{E}^\text{inc}_\text{DOI}$ denote the total and incident electric fields at an observation point $\mathbf{r}$. The background medium is assumed to be air and $g$ is the two-dimensional dyadic Green's function in free space. $\mathbf{J}$ is the induced current density. $k_0$ stands for the wave number of background medium. Eq.~\eqref{eq:stateEqu} describing the multiple scattering behavior inside DOI is also referred to as the state equation \cite{chen2018computational}. 

The induced current density has a nonlinear relation with contrast $\chi$,
\begin{equation}
	\mathbf{J}(\mathbf{r}^\prime) = {\chi}(\mathbf{r}^\prime)\mathbf{E}^\text{tot}_\text{DOI}(\mathbf{r}^\prime),
	\label{eq:nonlinearEqu}
\end{equation}
where ${\chi} = {\epsilon_r}-1$, $\epsilon_r$ denotes the relative permittivity. The non-linearity comes from the presence of the induced current distribution in the state equation ~\eqref{eq:stateEqu}, which is due to the phenomenon of multiple scattering. Then, the far fields generated by the induced currents are computed using the so-called data equation
\begin{equation}
	\mathbf{E}^\text{sca}_\text{mea}(\mathbf{r}) = k_0^2\int_\text{DOI}g(\mathbf{r},\mathbf{r}^\prime)\mathbf{J}(\mathbf{r}^\prime)d\mathbf{r}^\prime, 
	\label{eq:dataEqu}
\end{equation} 
where $\mathbf{E}^\text{sca}_\text{mea}$ denotes scattered field measured by receivers. Multiple input multiple output measurement scenario is assumed so that the DOI is illuminated by different transmitters, and multiple receivers collect the scattered electric far-fields.

Inverse scattering problems aim at retrieving the distribution of the relative permittivity within DOI based on the collected $\mathbf{E}^\text{sca}_\text{mea}$ and the knowledge of Green's function $g$. 
Regularization \cite{oliveri2017compressive} is often used to reduce the ill-posedness by imposing prior information on the desired solution of contrast function, e.g., sparse representation in the space supported by specific basis. Together with the constraints on data discrepancies, $\boldsymbol{\chi}$ is obtained by solving the following optimization problem
\begin{equation}
	\min_{\hat{\boldsymbol{\chi}}} L({\hat{\boldsymbol{\chi}}}) =  ||\mathbf{E}^\text{sca}_\text{mea}-\hat{\mathbf{E}}^\text{sca}_\text{mea}({\hat{\boldsymbol{\chi}}})||_2^2 + \beta \mathcal{R}({\hat{\boldsymbol{\chi}}}),
	\label{eq:optimizationEq}
\end{equation} 
where $\hat{\mathbf{E}}^\text{sca}_\text{mea}$ is the computed scattered field assuming a candidate contrast distribution ${\hat{\boldsymbol{\chi}}}$.
$\mathcal{R}({\hat{\boldsymbol{\chi}}})$ is the regularization term to impose the prior information on $\hat{\boldsymbol{\chi}}$ and $\beta$ is the regularization parameter trading off the contribution from data-fitting error and regularization cost. The solution to \eqref{eq:optimizationEq} is obtained following iterative algorithms. Two challenges exist to solve the optimization solver, the proper selection of regularizer and the value of $\beta$. Both are still open questions. While we refrain from detailing the different solutions explored in the past, we do mention that popular solutions include using the state equation (Eq. \eqref{eq:stateEqu}, \cite{chen2009subspace}), induced currents (Eq. \eqref{eq:nonlinearEqu}, \cite{van1997contrast}), contrast modification for non-linearity reduction \cite{chew1990reconstruction,agarwal2012improving}, sparsity prior \cite{oliveri2017compressive}, etc. Also, regularizations through alternate bases, such as eigenbases \cite{agarwal2012improving}, Fourier bases \cite{zhong2011fft} and level sets \cite{dorn2006level}, have been explored.  We refer to all of them collectively as 'traditional methods'.

\section{Deep neural network solvers}
\label{sec:DNN}

While applying DNNs to solve the ISPs, the Euclidean loss of the solution error is often taken as the loss function during the training process. Following the analogy between the DNNs and conventional iterative algorithms, one finds that the physical information of the imaging system is missing in the loss function. The role of physical information through the use of data equation ends in providing the initial estimate and DNN itself does not use either the data equation or the near field priors. This analogy is elaborated in Section \ref{subsec:analogy}. In order to rectify this blindness of the DNN to the scattering physics, two physics-guided loss functions are proposed in Section \ref{subsec:newlossfunctions}.

\subsection{Analogy between neutral network and iterative algorithm}
\label{subsec:analogy}

It has been shown that layered fully-connected neutral networks resemble iterative solvers \cite{li2018deepnis}. Taking Born iterative method (BIM) \cite{wang1989iterative} for instance, the solver follows a recursive scheme based on the matrix form of \eqref{eq:stateEqu}, 
\begin{equation}
	\mathbf{E}^\text{tot}_{\text{DOI},(p)} = \mathbf{E}^\text{inc}_\text{DOI} + ak_0^2\mathbf{g}_\text{DOI}\text{diag}(\mathbf{E}^\text{tot}_{\text{DOI},(p-1)})\boldsymbol{\chi}_{(p-1)},
\end{equation}
where $a$ denotes the area of a single subunit in the DOI (assuming subunits are with the same size), the subscript $(p)$ stands for the solution at the $p$th iteration, and $\mathbf{g}_\text{DOI}$ denotes the Green's function when the observation points are within DOI. With the updated $\mathbf{E}^\text{tot}_{\text{DOI},(p)}$ and following the formulation of \eqref{eq:optimizationEq}, $\boldsymbol{\chi}_{(p)}$ is the solution to the optimization problem  
\begin{equation}
	\boldsymbol{\chi}_{(p)} = \arg\min_{\hat{\boldsymbol{\chi}}} \frac{1}{2}||\mathbf{E}^\text{sca}_\text{mea} - \mathbf{G}_{(p)}\hat{\boldsymbol{\chi}}||^2_2 + \beta ||\hat{\boldsymbol{\chi}}||_1,
	\label{eq:exampleOptimizationEq}
\end{equation}
where $\mathbf{G}_{(p)} = ak_0^2\mathbf{g}_\text{mea}\text{diag}(\mathbf{E}^\text{tot}_{\text{DOI},(p)})$ and $\mathbf{g}_\text{mea}$ is the Green's function when the observation points locate at receiver positions. The recursive schemes starts from $\mathbf{E}^\text{tot}_{\text{DOI},(0)} = \mathbf{E}^\text{inc}_\text{DOI}$. The regularization term assumes the contrast function is sparse itself and the sparsity is quantified by its $l_1$ norm. The iterative shrinkage and thresholding algorithm (ISTA) \cite{daubechies2004iterative} gives the recursive solution of \eqref{eq:exampleOptimizationEq} as
\begin{equation}
	\boldsymbol{\chi}_{(p,q)} = \mathcal{S}_{\beta/L}\left\lbrace \mathbf{A}_{(p,q-1)}\boldsymbol{\chi}_{(p,q-1)}+\mathbf{b}_{(p,q-1)}\right\rbrace, 
	\label{eq:recursiveEq}
\end{equation}
where $\boldsymbol{\chi}_{(p,q)}$ denotes the solution of $\boldsymbol{\chi}_{(p)}$ at the $q$th iteration. $\mathcal{S}_{\beta/L}$ is a soft-thresholding operator 
and $L$ is the Lipschitz constant of a normal operator and should be smaller than eigenvalues of $\mathbf{G}^H_{(p)}\mathbf{G}_{(p)}$ \cite{jin2017deep}, where the superscript $H$ indicates Hermitian. $\mathbf{A}_{(p,q-1)}$ and $\mathbf{b}_{(p,q-1)}$ have expressions
\begin{subequations}
	\label{eqs:defWeights}
	\begin{equation}
		\mathbf{A}_{(p,q-1)}=\mathbf{I}-\frac{1}{L}\mathbf{G}^H_{(p)}\mathbf{G}_{(p)},
	\end{equation}
	\begin{equation}
		\mathbf{b}_{(p,q-1)} = \frac{1}{L}\mathbf{G}^H_{(p)}\mathbf{E}^\text{sca}_\text{mea},
	\end{equation}
\end{subequations}
where $\mathbf{I}$ is the identity matrix. From \eqref{eq:recursiveEq}, one sees that $\mathbf{A}_{(p,q-1)}$ and $\mathbf{b}_{(p,q-1)}$ can be interpreted as the weights and the biases for nodes at the $(q-1)$th layer of the $p$th module in a hypothetical neural network designed to emulate the iterative process of BIM. Further, $\mathcal{S}_{\beta/L}$ is a nonlinear operator defined as
\begin{equation}
	S_\theta = \left\lbrace 
	\begin{array}{ll}
		x+\theta/2, & \text{if}\,\, x\le -\theta/2,\\
		0, & \text{if}\,\, |x|<\theta/2,\\
		x-\theta/2, & \text{if}\,\, x\ge \theta/2.
	\end{array}\right.
\end{equation}
It is easy to see that $\mathcal{S}_{\beta/L}$ acts as the activation function in the neurons of such hypothetical neural network.

This analogy provides ways to intuitively understand the DNN learning process via the analysis of iterative algorithms, such as BIM above. The input of the network plays the role of initial solution $\boldsymbol{\chi}_{(0,0)}$ in this iterative scheme. From the expressions of weights and biases in \eqref{eqs:defWeights}, we expect that the following iterative procedures needs the knowledge of Green's function and the measured scattered fields, which, however, do not participate into the training process of DNN solvers. This indicates that although an analogy may be derived such as above, the actual DNN solvers for solving ISPs do not correlate to the above analogy and their own neural networks do not incorporate the physical aspects in their weights and biases, such as described in Eq.~\eqref{eqs:defWeights}. As discussed in the next sub-section, this insight paved the path to our proposed physics-guided loss functions.


\subsection{Physics-guided loss functions} \label{subsec:newlossfunctions}

The DNN solvers sketched in Fig.\ref{fig:sketchISP} actually treats the input and the response as images. DNN solvers of ISPs conventionally use the following as the loss function
\begin{equation}
	L^{\text{contrast}}(\hat{\boldsymbol{\chi}}) =||\hat{\boldsymbol{\chi}}-\boldsymbol{\chi}||_2^2,
	\label{eq:reg_loss}
\end{equation}
with which the trained network can effectively learn patterns of the scattering contrast distributions in the training database so that the response fits the learned patterns. However, in the traditional ISPs, the scattering contrast discrepancy functions as a regularizer. For example, applying this regularizer in \eqref{eq:optimizationEq}, the optimization problem becomes 
\begin{equation}
	\boldsymbol{\chi}_{(p)} = \arg\min_{\hat{\boldsymbol{\chi}}} \frac{1}{2}||\mathbf{E}^\text{sca}_\text{mea} - \mathbf{G}_{(p)}\hat{\boldsymbol{\chi}}||^2_2 + \beta ||\hat{\boldsymbol{\chi}}-\boldsymbol{\chi}||_2^2.
	\label{eq:optimizationEqDL}
\end{equation} 

It is worth considering if the minimization function above can be used as a physics-guided loss function for DNNs. In the noiseless cases, the data-fitting term may not contribute much since the initial estimate provided as input to the DNN is already close to the ground truth and the data-fitting error is small to begin with. However, when the SNR is poor, considering data-fitting errors in the loss function may be helpful for improving the prediction robustness. However, since receivers are positioned far from the DOI, contributions from evanescent waves that encode multiple scattering cannot be collected. The missing information leads to ill-conditioned matrix $\mathbf{G}_{(p)}$. Here, the state equation, instead of data equation, is used to constrain the data-fitting error. Since the field solution of DOI can be computed based on $\boldsymbol{\chi}$ and $\mathbf{E}^\text{inc}_\text{DOI}$,
we have
\begin{equation}
	\boldsymbol{\chi}_{(p)} = \arg\min_{\hat{\boldsymbol{\chi}}} \frac{1}{2}||\mathbf{E}^\text{sca}_\text{DOI} - \mathbf{G}_\text{DOI}^{(p)}\hat{\boldsymbol{\chi}}||^2_2 + \beta ||\hat{\boldsymbol{\chi}}-\boldsymbol{\chi}||_2^2.
	\label{eq:optimizationStateEqu}
\end{equation}
where $\mathbf{E}^\text{sca}_\text{DOI}=\mathbf{E}^\text{tot}_\text{DOI}-\mathbf{E}^\text{inc}_\text{DOI}$. $\mathbf{G}_\text{DOI}^{(p)}$ indicates multiple scattering effects in DOI and is less ill-conditioned due to small distances between observation points and scatterers. Remark that to simulate noisy measurements, $\mathbf{E}^\text{sca}_\text{DOI}$ is artificially corrupted by Gaussian noise. 
By incorporating the scattered near-field in the loss function, both physical information and noise effects are therefore considered in the training process, improving the robustness of DNN. 

Following similar arguments about incorporating near-field priors, feature-enhanced imaging is obtained by designing the constraint about the induced current, i.e.,
\begin{equation}
	\boldsymbol{\chi}_{(p)} = \arg\min_{\hat{\boldsymbol{\chi}}} \frac{1}{2}||\mathbf{J} - \mathbf{E}^\text{tot}_\text{DOI}\odot\hat{\boldsymbol{\chi}}||^2_2 + \beta ||\hat{\boldsymbol{\chi}}-\boldsymbol{\chi}||_2^2,
	\label{eq:optimizationEq_enhancedFeature}
\end{equation}
where $\odot$ denotes the operator of element-wise product. Since the positions of nonzero elements of $\mathbf{J}$ imply the geometry of $\boldsymbol{\chi}$, the constraint about $\mathbf{J}$ could enhance the feature learning.  

\begin{figure*}[!th]
	\includegraphics[width=\linewidth]{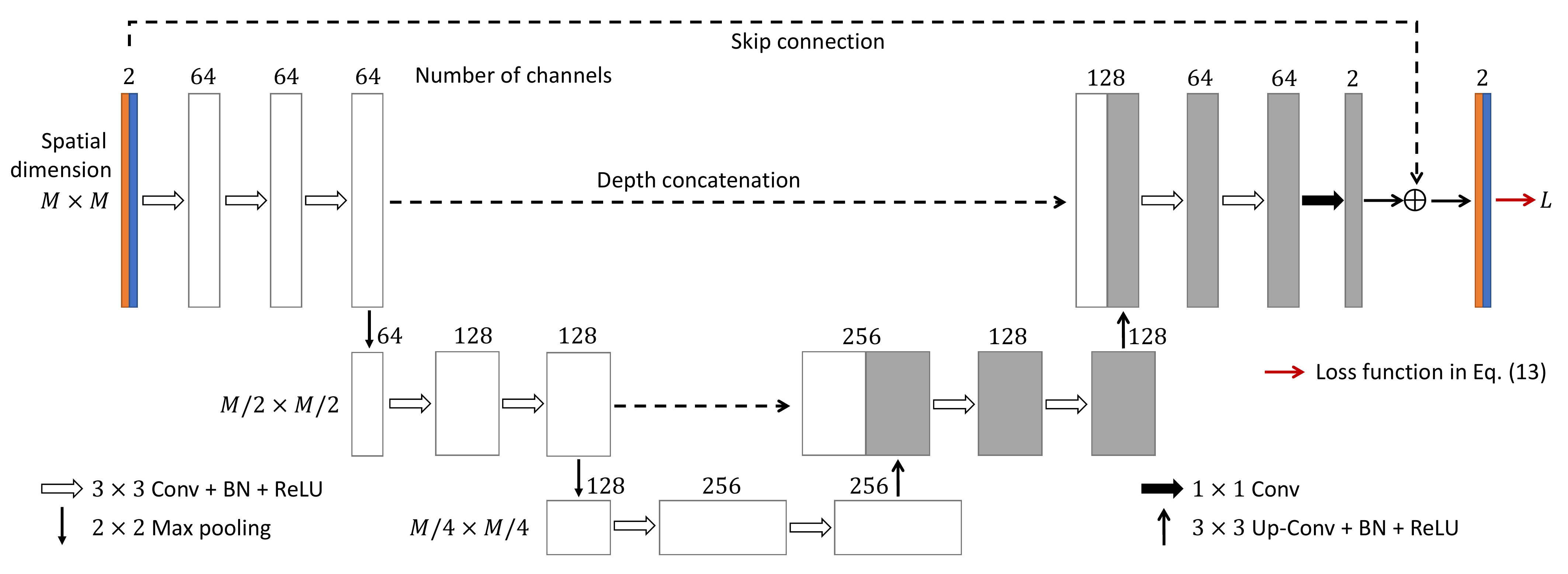}
	\caption{The U-net architecture used to train a inverse scattering problem solver. The input has two channels for real part and imaginary part of initial estimation of contrast. BN is short for batch normalization. Details can be found in \cite{ronneberger2015u} and \cite{jin2017deep}.}
	\label{fig:UNetArchitecture}
\end{figure*}

\section{Numerical results}
\label{sec:results}

\subsection{Training and test datasets} 
The imaging performances with different loss functions are studied based on simulation results. With wave illuminations at a single frequency, the scattered electric far-field solutions of  digit-like and polygon-like scatterers are obtained using method of moment with the pulse basis function and the delta test function to discretize the DOI of size $5.6\lambda_0\times 5.6\lambda_0$ into $64\times 64$ pixels, $\lambda_0 = \SI{7.5}{\centi\meter}$. Noisy measurement are obtained by adding Gaussian noise to the scattered fields with a specific signal-to-noise ratio.
The measurement scenario is described next. $36$ line sources sequentially emit waves and the same number of line receivers are used to collect the scattered fields. The sources and receivers are uniformly placed on the circle, which shares the center with DOI and has a radius $10\lambda_0$. 

The training set is computed based on the MNIST database \cite{deng2012mnist}, which is composed of images of handwritten digits. The training set includes $1000$ images and an independent set of $2000$ images is used for testing. The virtual scatterers are generated by setting pixels with values below the one third of the maximum as air background and the remaining part as a homogeneous dielectric scatterer with a constant relative permittivity. The relative permittivity is assigned a random value in the range $[1,5]$ in order to include scatterers of varying challenges in terms of the non-linearity arising from high contrast \cite{agarwal2012improving}. 

To test the generalizability, we also designed a different dataset, which consists of polygonal shapes that are quite different from the shapes of digits. $2000$ virtual polygon-like scatterers are designed by randomly positioning regular polygons and setting the relative permittivity in the range $[1, 5]$. The number of sides varies from $3$ to $7$ and the distance between vertices to the geometric center ranges from $0.1\lambda_0$ to $1.6\lambda_0$. Overlapping between polygons is allowed.

\subsection{Network settings}
\label{subsec:networkSetting}

U-net \cite{ronneberger2015u} is a DNN that has been widely applied in deep-learning-based image translation algorithms. It has shown advantages in prediction accuracy compared with traditional ISP solvers. Here, we apply the U-net architecture adapted from \cite{wei2018deep} and described in Fig.~\ref{fig:UNetArchitecture}. The network is composed of two parts, contraction path (encoder in the parlance of deep learning) and expansive path (decoder). Both paths include repeated convolution layers comprising of kernels of size $3\times3$ pixels, batch normalizations, and rectified linear unit (ReLU). At each down-sampling step in the contraction path, the spatial dimension is halved through a $2\times 2$ max pooling operation and the number of channels is doubled by doubling the number of kernels in the next convolution layer. The up-sampling steps follows the contrary procedures except that after doubling the spatial dimension using $3\times3$ up-convolutions, batch normalization and ReLU activation are performed. In the U-net architecture, the down-sampling path encodes the input information and the up-sampling path decodes the information to reconstruct the contrast profile. Skip connections in U-Net support the learning of the residual between the input contrast function and the ground truth.

The input to U-net is an initial estimate of contrast function generated by the fast conventional ISP solver back propagation (BP) \cite{wei2018deep}. In addition, the architecture, the loss function, and the conditions of \cite{wei2018deep} are replicated as the relevant state-of-the-art for comparison of the performance of our approaches. Since scatterers in detection could be lossy, complex inputs have been considered by treating the real part and the imaginary part as two independent input channels such that this complex-valued problem can be solved using the real-valued neutral network.

\subsection{Training, testing, and performance evaluation}
We compare three candidate loss functions for learning the network presented below, of which the first one $L^{\text{contrast}}$ has been used in \cite{wei2018deep} and the other two $L^{\text{current}}$ and $L^{\text{field}}$ are the proposed physics-guided loss functions.
\begin{subequations}
	\label{eq:threeLossFunctions}
	\begin{align}
		\begin{split}
			L^{\text{contrast}}(\hat{\boldsymbol{\chi}}) &= ||\boldsymbol{\chi}-\hat{\boldsymbol{\chi}}||_2^2,
			\label{eq:lossContrast}
		\end{split}\\
		\begin{split}
			L^{\text{current}}(\hat{\boldsymbol{\chi}}) &= \frac{1}{2}||\mathbf{J} - \mathbf{E}^\text{tot}_\text{DOI}\odot\hat{\boldsymbol{\chi}}||^2_2 + \beta ||\boldsymbol{\chi}-\hat{\boldsymbol{\chi}}||_2^2,
			\label{eq:lossCurrent}
		\end{split}\\
		\begin{split}
			L^{\text{field}}(\hat{\boldsymbol{\chi}}) &= \frac{1}{2}||\mathbf{E}^\text{sca}_\text{DOI} - \mathbf{G}_\text{DOI}\hat{\boldsymbol{\chi}}||^2_2 + \beta ||\boldsymbol{\chi}-\hat{\boldsymbol{\chi}}||_2^2.
			\label{eq:lossField}
		\end{split}
	\end{align}
\end{subequations}
Benefiting from the available true solution of all involved quantities, the regularization parameter $\beta$ is set as $2||\mathbf{J}||^2_2/||\boldsymbol{\chi}||_2^2$ for $L^{\text{current}}$ and  $2||\mathbf{E}^\text{sca}_\text{DOI}||^2_2/||\boldsymbol{\chi}||_2^2$ for $L^{\text{field}}$, respectively. The physical quantities in $\beta$ make the two terms comparable in terms of units and scale. The factor $2$ gives more weight to the scattering contrast discrepancies and the additional near-field constraints about induced current and scattered fields provide supplementary information to refine the estimation. We note that the normalization parameter $\beta$ in the proposed loss functions is computed for one batch of images for which the loss is computed collectively by the deep learning training algorithm.

As mentioned in Section \ref{sec:DNN}, noisy scenarios are simulated by adding \SI{5}{dB} Gaussian noise (except studies in Section \ref{subsec:SNR-Effects} and \ref{subsec:imagingCylinders}) to $\mathbf{E}^\text{sca}_\text{DOI}$. Same SNR is applied to $\mathbf{E}^\text{sca}_\text{mea}$ when noisy input is considered. To make fair comparisons, both noisy input and clean input with the loss function $L^{\text{contrast}}$ are analyzed here. In the following analysis, notations $L^{\text{contrast}}_\text{clean}$ and $L^{\text{contrast}}_\text{noisy}$ are used to distinguish the corresponding results. 

The deep learning toolbox in MATLAB 2020a is used to implement the training method. The same configurations for training have been used for the different loss functions. Stochastic gradient descent with momentum $0.99$ is used to optimize weights and biases of the network. Learning rate is set at $5\times 10^{-6}$ in the beginning and is halved every $20$ epochs. The learning process is stopped when the max of epochs $150$ is reached. A computational server is used for the training process and the learnt model is transferred to a laptop with CPU ($\SI{2.6}{\giga\hertz}$ processor and $\SI{16}{\giga\byte}$ memory) for prediction and evaluation on test datasets. The performance of the trained network is evaluated by computing mean square error (MSE) and structural similarity index measure (SSIM) between the predicted and the ground truth contrast profiles. In the following, the results on test datasets are presented. 

Imaging performances of DNN solvers with different loss functions are studied considering two different values of SNRs. 
To make comparisons with conventional methods, Born iterative method (BIM) \cite{wang1989iterative} is used. The input to the neural network solvers, i.e. the results of back propagation (BP) are also presented. For the ease of reference, we call each trained network a `solver', in the same vein as BIM and BP solvers. We refer to each solver by either its acronym (BIM, BP) or the loss function. DNN solvers are obtained with the three loss functions in \eqref{eq:threeLossFunctions}, while the subscripts of $L_\text{clean}^\text{contrast}$ and $L_\text{noisy}^\text{contrast}$ stand for clean and noisy input, respectively.

\subsection{Results for digit-like scatterers similar to training set}

\begin{figure}[!ht]
	\centering
	\includegraphics[width=\linewidth]{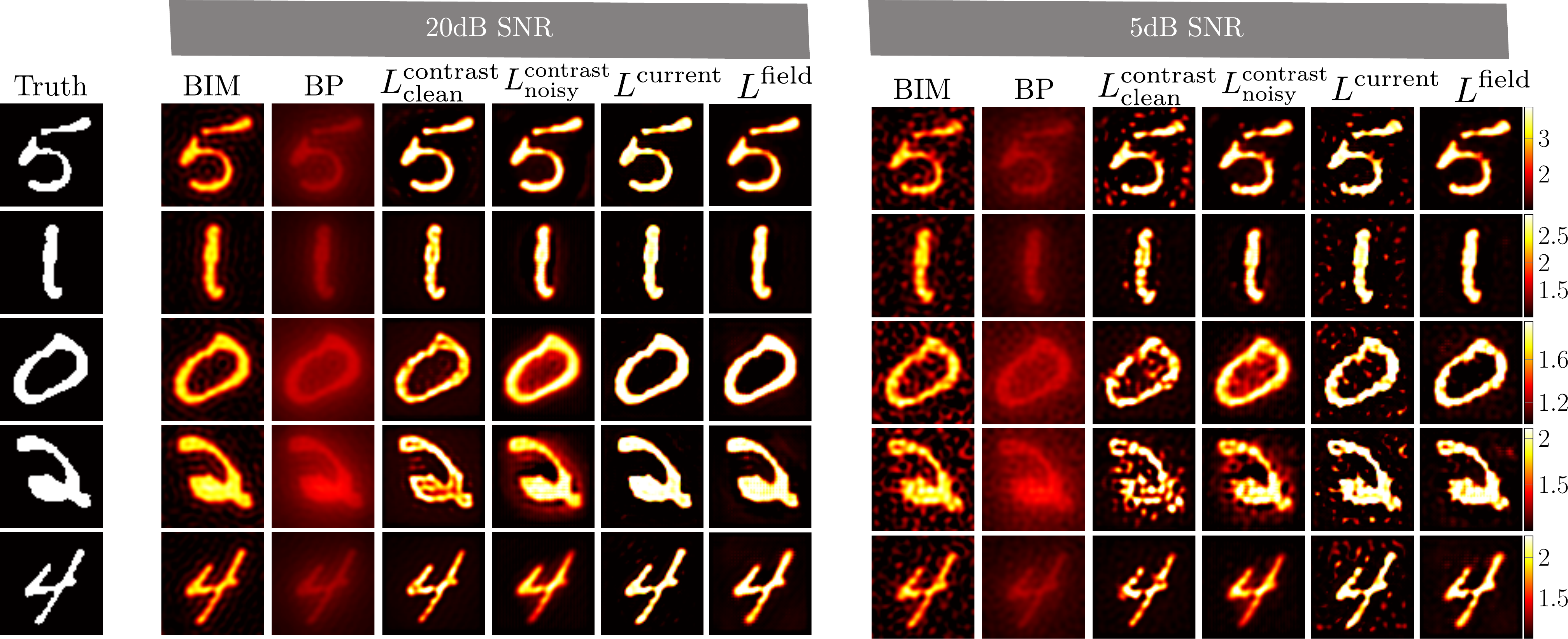}
	\caption{Testing results on digit-like scatterers when measured scattered fields are corrupted by Gaussian noises with SNR = \SI{20}{\dB} and \SI{5}{\dB}. Born iterative method (BIM), back propagation (BP) and DNN ISP solvers trained based on MNIST database are applied. Results corresponding to $L^\text{contrast}_\text{clean}$ are with clean input and the loss function only penalizing the contrast discrepancies, while $L^\text{contrast}_\text{noisy}$ denotes results with noisy input. By additionally imposing constraints about induced current and scattered fields with DOI, the results are labeled by $L^\text{current}$ and $L^\text{field}$. } 
	\label{fig:resultsMNIST}
\end{figure}

For digit-like scatterers, five representative examples are given in Fig.~\ref{fig:resultsMNIST}. Images from different solvers are shown with the same color range for easier visual comparisons. With \SI{20}{\dB} SNR of measured scattered fields $E_\text{mea}^\text{sca}$, the results are in the left panel. While the shape of the scatterers are well reconstructed with all the given methods, BIM and BP underestimate the relative permittivities of scatterers. Since the SNR is high, when only penalizing contrast discrepancies, the network trained from clean input, i.e., with $L_\text{clean}^\text{contrast}$, fits the testing scenarios more than the cases with $L_\text{noisy}^\text{contrast}$, the results of which suffer from the shadow-like or halo-like effects. On the other hand, the result of $L_\text{clean}^\text{contrast}$ suffers from small background debris artifacts especially notable in the third and the fourth example, which are seen also in BIM and are present in extremely low contrast in BP too. The halo and debris artifacts are suppressed by additionally considering data-fitting errors from near scattered fields, i.e., with $L^\text{field}$. However, the DNN solver with $L^\text{current}$ is obviously more capable of obtaining high-quality imaging results.

When the SNR is reduced to \SI{5}{\dB}, although the shape information of digits is well reconstructed by all the solvers, but the obtained images are imperfect in different ways. For BIM and BP, relative permittivities are still underestimated and artifacts are uniformly distributed in the whole image. Among $L^{\text{contrast}}_{\textbf{clean}}$ and $L^{\text{contrast}}_{\textbf{noisy}}$, $L^{\text{contrast}}_{\textbf{noisy}}$ has smaller Euclidean cost with the ground truth. As a result, the estimation of the relative permittivity of scatterers is less biased than the results of BIM, BP and $L^{\text{contrast}}_{\textbf{clean}}$, and the artifacts are more sparsely distributed. Compared with $L^{\text{contrast}}_{\textbf{clean}}$, the DNN solver trained with $L^{\text{contrast}}_{\textbf{noisy}}$ learns to compensate for noise, which results in scarcer artifacts in the background present mainly around the scatterers. After applying the constraint related with induced current, i.e. training using $L^{\text{current}}$, the accuracy of estimated relative permittivity of scatterers is significantly improved but at the cost of increased background artifacts, which on the other hand are successfully suppressed using $L^{\text{field}}$. 

\begin{sidewaystable}
	\caption{Statistical parameters for test results of digit-like and polygon-like scatterers. Legend: \textbf{Best}, \underline{second best}.}
	\label{tab:statisticalResults}
	\centering
	\subfloat[Mean square error (MSE)]{
		\begin{tabular}{ll|cccc|cccc|cccc}
			\toprule
			&& \multicolumn{4}{c|}{Mean} & \multicolumn{4}{c|}{Median} & \multicolumn{4}{c}{Standard deviation}\\ 
			\midrule
			SNR & Dataset & $L^\text{constrast}_\text{clean}$ & $L^\text{constrast}_\text{noisy}$ & $L^\text{current}$ & $L^\text{field}$ & $L^\text{constrast}_\text{clean}$ & $L^\text{constrast}_\text{noisy}$ & $L^\text{current}$ & $L^\text{field}$& $L^\text{constrast}_\text{clean}$ & $L^\text{constrast}_\text{noisy}$ & $L^\text{current}$ & $L^\text{field}$\\
			\midrule
			\midrule
			20 dB &	Digit & 0.0715& 0.0817& \textbf{0.0404} & \underline{0.0701} & 0.0538& 0.0610& \textbf{0.0277} & \underline{0.0460} & 0.0530& 0.0614& 0.0382& 0.0625\\
			& Polygon & 0.1147& 0.0917& \underline{0.0716} & \textbf{0.0625}& 0.0352& 0.0382& \textbf{0.0249} & \underline{0.0333} & 0.2848& 0.1874& 0.1577& 0.1010\\
			\midrule
			5 dB & Digit & 0.1475& \underline{0.1015} & 0.1156& \textbf{0.0888}& 0.0963& 0.0727& \underline{0.0699}& \textbf{0.0601}& 0.1496& 0.0849& 0.1434& 0.0815\\
			& Polygon & 0.2088& \underline{0.1167}& 0.1674& \textbf{0.0836}& 0.0469& \underline{0.0428}& 0.0545& \textbf{0.0399}& 0.4609& 0.2480& 0.3515& 0.1428\\
			\bottomrule		
	\end{tabular}}\\
	\subfloat[Structural similarity index (SSIM)]{
		\begin{tabular}{ll|cccc|cccc|cccc}
			\toprule
			& & \multicolumn{4}{c|}{Mean} & \multicolumn{4}{c|}{Median} & \multicolumn{4}{c}{Standard deviation}\\ 
			\midrule
			SNR & Dataset & $L^\text{constrast}_\text{clean}$ & $L^\text{constrast}_\text{noisy}$ & $L^\text{current}$ & $L^\text{field}$ & $L^\text{constrast}_\text{clean}$ & $L^\text{constrast}_\text{noisy}$ & $L^\text{current}$ & $L^\text{field}$& $L^\text{constrast}_\text{clean}$ & $L^\text{constrast}_\text{noisy}$ & $L^\text{current}$ & $L^\text{field}$\\
			\midrule
			20 dB & Digit & 0.6252& 0.6066& \textbf{0.8071}& \underline{0.6694}& 0.6222& 0.6015& \textbf{0.8373}& \underline{0.6807}& 0.0828& 0.0563& 0.0849& 0.0800\\
			& Polygon & \underline{0.6627} & 0.6315& \textbf{0.6742}& 0.6378& \textbf{0.7255}& 0.6813& 0.\underline{6956}& 0.6368& 0.1578& 0.1302& 0.1634& 0.0906\\
			\midrule
			5 dB & Digit & 0.4696& \underline{0.5298}& 0.5025& \textbf{0.5703}& 0.4595& \underline{0.5160}& 0.4844& \textbf{0.5748}& 0.1523& 0.0857& 0.1392& 0.0662\\
			&	Polygon & 0.5335& \textbf{0.5541}& 0.3961& \underline{0.5497}& \textbf{0.6361}& \underline{0.6253}& 0.3281& 0.5578& 0.2449& 0.1876& 0.2190& 0.1555\\
			\bottomrule
	\end{tabular}}
\end{sidewaystable}

Statistical analysis of performances of the DNN solvers are performed based on 2000 digit-like test examples. MSE and SSIM are used for this comparison, where accurate solution has MSE equal to zero and SSIM equal to 1. Their mean, median and standard deviation are presented in Table~\ref{tab:statisticalResults}. The superiority of the DNN solver with $L^{\text{current}}$ in high SNR scenarios is observed. In terms of MSE, $L^{\text{current}}$ is shown as the best option from the comparison of mean, median and standard deviation. When quantified by SSIM, although the standard deviation is slightly larger than the other solvers, the significant advantage of the mean value still support the superiority of the DNN solver with $L^{\text{current}}$. In low SNR scenarios, however, $L^{\text{current}}$ only has advantages to $L_\text{clean}^\text{contrast}$ and the prediction performance degrades heavily. Since noise effects are considered in input and the loss function, the solver with $L^\text{field}$ is superior to the other solvers in the noise robustness. Moreover, the comparison of imaging performances between \SI{20}{\dB} and \SI{5}{\dB} reveals that the solver with $L^\text{current}$ is much more sensitive to the variation of noise level than $L^\text{field}$.

\subsection{Results on polygon-like structures not used in training}

\begin{figure}[!t]
	\centering
	\includegraphics[width=\linewidth]{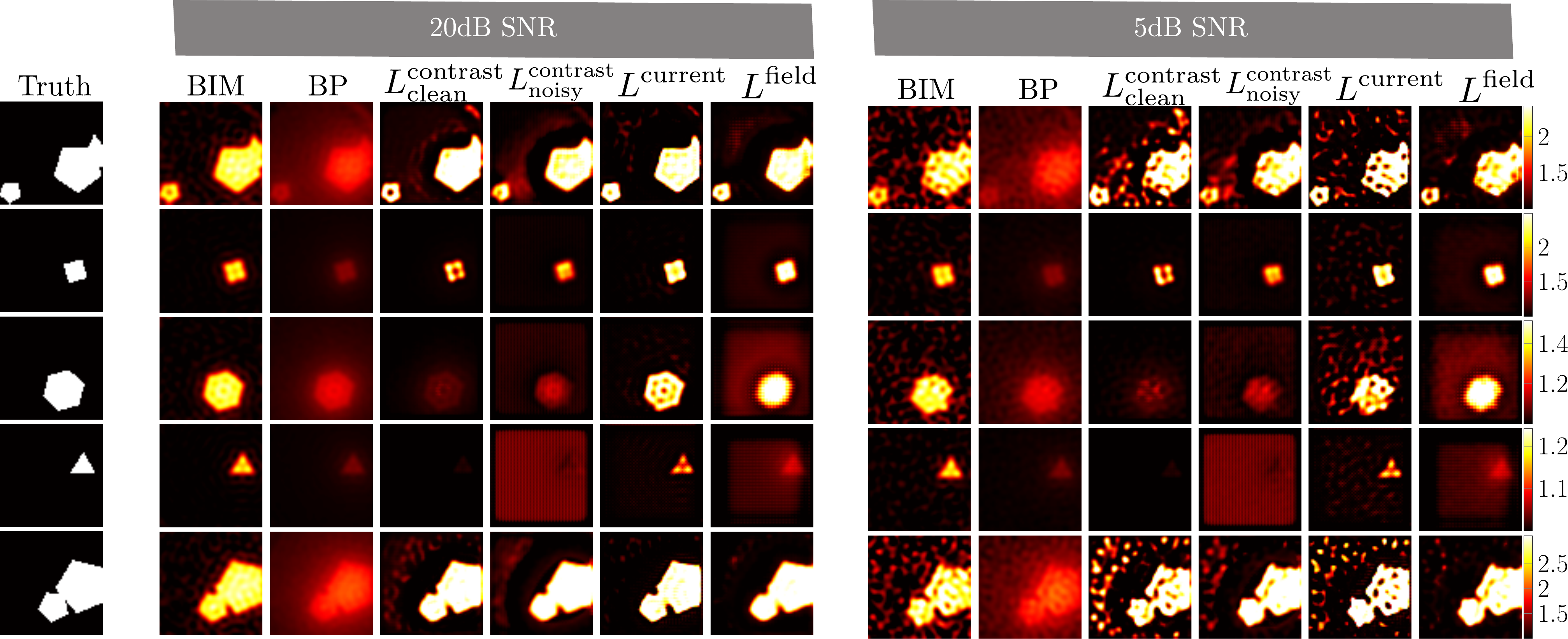}
	\caption{Testing results on polygon-like scatterers when measured scattered fields are corrupted by Gaussian noises with SNR = \SI{20}{\dB} and \SI{5}{\dB}. Five representative examples and statistical evaluation results based on $2000$ testing samples are  presented.}
	\label{fig:resultsPolygon}
\end{figure}

With the same DNN solvers, which are trained using the MNIST database, test results on polygon-like scatterers are shown in Fig.~\ref{fig:resultsPolygon}. The conclusions made from the observations with digit-like scatterers are also applicable here in addition to the new finding that scatterers with small contrast are poorly predicted by all the DNN solvers.

As seen from the fourth example, both of them fail when the contrast is small. Improvements are observed with $L^\text{field}$, although BIM and feature-enhanced DNN solver with $L^\text{current}$ perform better in this case. Comparisons of MSE and SSIM between digit- and polygon-like scatterers reveal that the additional constraint about induced current could make the imaging performance of DNN solvers more dependent on the training set. Scatterers similar with samples in the training set are more accurately imaged. Moreover, they fail in imaging low contrast scatterers when only penalizing contrast discrepancies. We also noted the dependence of the performance of DNN solver with $L^\text{current}$ on the training dataset.

\subsection{SNR of training data versus SNR of test data}
\label{subsec:SNR-Effects}

\begin{figure}[!ht]
	\centering
	\includegraphics[width=0.5\linewidth]{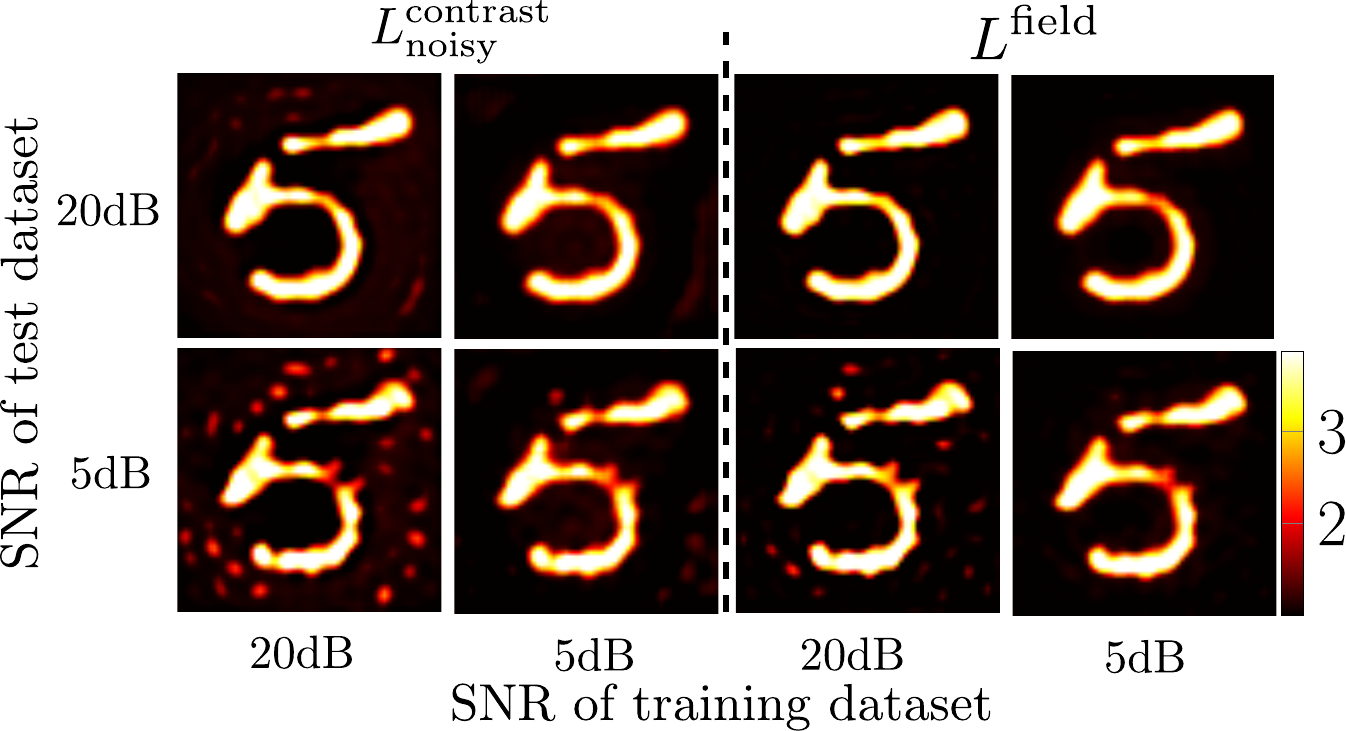}
	\caption{Effects of SNR mismatch between training and test.}
	\label{fig:SNR_Effects}
\end{figure}
The accurate value of SNR is unknown in real scenarios. Therefore, it is of interest to study the effect of mismatch in the SNR of the simulated training dataset and the SNR of test samples. Since the input for the DNN solvers with $L_\text{clean}^\text{contrast}$ and $L^\text{current}$ does not consider noise,  only example results of $L_\text{noisy}^\text{contrast}$ and $L^\text{field}$ are given in Fig.~\ref{fig:SNR_Effects}. As seen, when the network is trained based on the  dataset with \SI{20}{dB} SNR, background artifacts appear with the \SI{5}{dB} test example. However, the trained network is robust to the SNR mismatching when the applied SNR values of the training dataset and the test example are reversed. Moreover, the DNN solver with $L^\text{field}$ seems more capable to deal with the SNR mismatching problem than the solver with $L_\text{noisy}^\text{contrast}$.

\subsection{Austria profile and a study in the contrast of scatterers}
\label{subsec:imagingCylinders}

\begin{figure}[!ht]
	\centering
	\includegraphics[width=\linewidth]{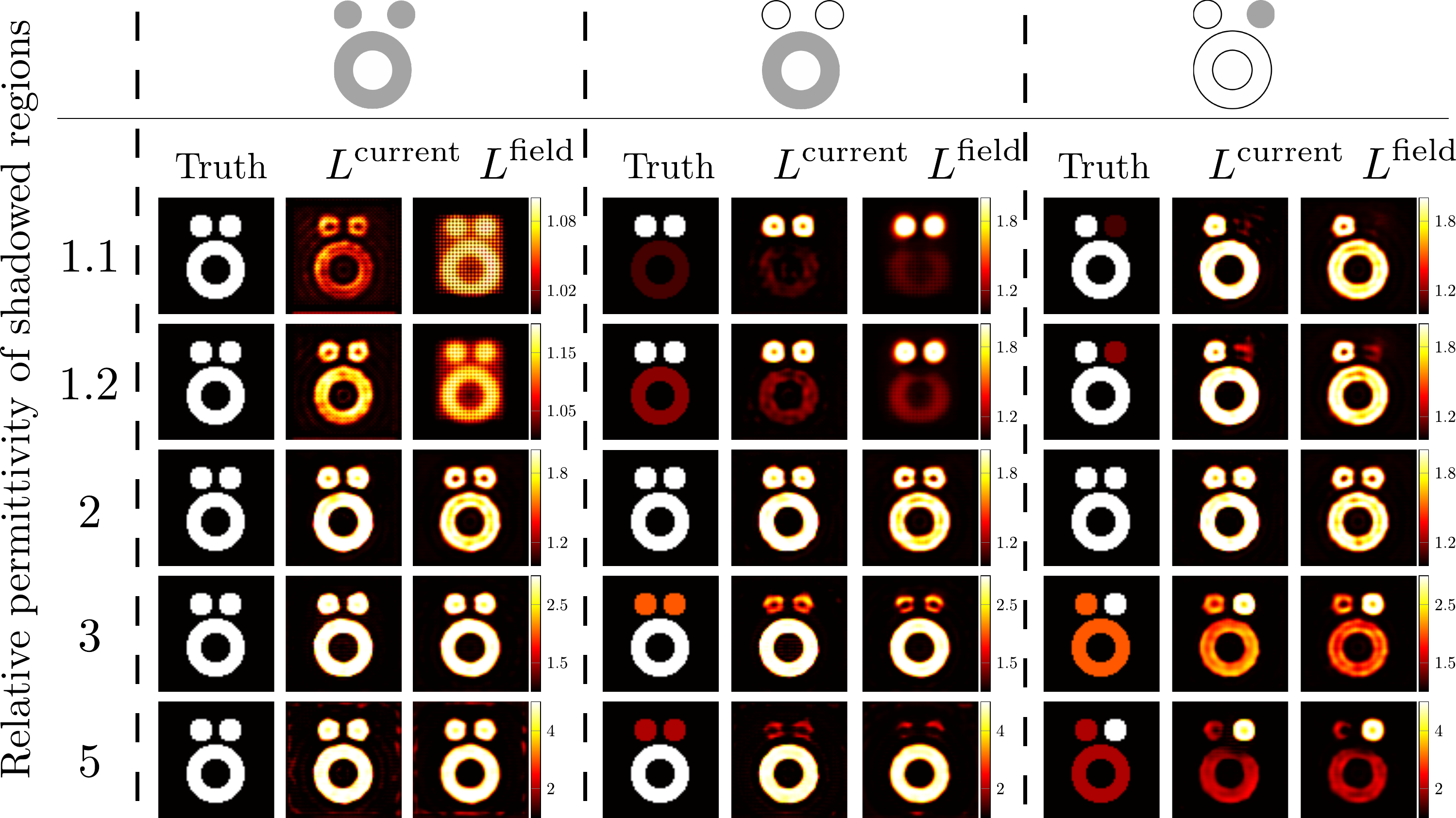}
	\caption{Imaging of Austria profile by the DNN solvers with loss functions $L^\text{current}$ and $L^\text{field}$ when the relative permittivity of the shadowed regions in the top panel varies and the remaining regions are with relative permittivity equal to $2$. SNR of training and test sets is \SI{20}{dB}.}
	\label{fig:Austria_variedPermittivity}
\end{figure}
The effect of variations of contrast on the performance of the proposed physics-guided loss functions are studied based on testing results on the representative and challenging ``Austria profile". The profile is made of two circular disks, which are with the same radius $0.56\lambda_0$ and centered at $[\pm 0.7\lambda_0, 1.4\lambda_0]$, and an annulus with interior radius $0.7\lambda_0$ and exterior radius $1.4\lambda_0$, respectively. All are made of homogeneous materials and the background is air. Three sets of variations are considered: (a) the relative permittivity of the entire profile is varied, (b) the relative permittivity of the annulus (i.e. the biggest scatterer) is varied while the relative permittivity of the other scatterers is fixed at $2$, and (c) the relative permittivity of one disk is varied while the relative permittivity of the other scatterers is fixed at $2$. For the scatterers whose relative permittivities are varied, the following candidates for relative permittivity values are considered $1.1$, $1.2$, $2$, $3$, to $5$. Here, $1.1$ and $1.2$ correspond to low contrast, $3$ is high contrast, while $5$ is considered very high contrast. The SNR of both the training set and the test samples is \SI{20}{\dB} SNR. 

The results are shown in Fig.~\ref{fig:Austria_variedPermittivity}. The set (a) simply studies the performance variation when the contrast of all the scatterers is the same but varies between low to high. It is seen that the loss functions are more effective for higher contrast. Further, the largest scatterer has more accurate reconstruction in general. The set (b) considers the variation of relative permittivity of the largest scatterer, i.e. the annulus. The scatterer(s) with high contrast are better reconstructed. However, the larger scatterer has a better shape reconstruction even when it has low contrast. On the other hand, the reconstruction of the shape of small scatterers is poorer than the large structure even when they have the same relative permittivity. The set (c) helps in comparing two scatterers of similar size when one has a different relative permittivity than the rest. Even in this situation, we find consistency in observation that the reconstruction is accurate in the order of higher contrast and larger size. 

In conclusion for this study, despite the arguments for concluding the superiority of $L^\text{current}$ to $L^\text{field}$ in high SNR scenarios, this serial study reveals that reconstructing low-contrast scatterers (relative to adjacent scatterers) is challenging for the two DNN solvers. When the entire profile is with the same permittivity, the position and the shape information of cylinders are well reconstructed. Otherwise, the parts with lower contrast can be poorly imaged and the cylinder geometry could be indistinguishable. This phenomenon is more evident for small scatterers.

\section{Summary of results and future outlook}
\begin{table}[!ht]
	\centering
	\caption{Qualitative evaluation of deep neural network solvers with different loss functions.}
	\label{tab:evaluation}
	\begin{tabular}{|m{2.2cm}|m{2cm}|m{2.8cm}|}
		\hline\vspace{3pt}
		\textcolor{white}{$L_\text{clean}^\text{contrast}$} 
		& \textbf{Best} & \textbf{Worst} \\
		\hline \hline \vspace{3pt}
		\textbf{Training time} & $L_\text{clean}^\text{contrast}$, $L_\text{noisy}^\text{contrast}$ & $L^\text{current}$, $L^\text{field}$\\[1pt]
		\hline
		\textbf{Accuracy (MSE)} & \vspace{2pt}$L^\text{field}$ (\SI{5}{\dB})\par\vspace{3pt} $L^\text{current}$ (\SI{20}{\dB}) &
		\vspace{2pt}$L_\text{clean}^\text{contrast}$ (\SI{5}{\dB})\par\vspace{3pt} $L_\text{noisy}^\text{contrast}$ (\SI{20}{\dB})\\[1pt]
		\hline
		\textbf{Accuracy (SSIM)} & $L^\text{field}$ (\SI{5}{\dB})\par\vspace{3pt} $L^\text{current}$ (\SI{20}{\dB}) & \vspace{2pt}$L_\text{clean}^\text{contrast}$ (\SI{5}{\dB}, digit) \par\vspace{3pt} $L^\text{current}$ (\SI{5}{\dB}, polygon) \par\vspace{3pt}  $L_\text{noisy}^\text{contrast}$ (\SI{20}{\dB})\\[1pt]
		\hline
		\textbf{Noise robustness} &\vspace{2pt} $L^\text{field}$ &\vspace{2pt} $L_\text{clean}^\text{contrast}$, $L^\text{current}$\\[1pt]
		\hline\vspace{3pt}
		\textbf{Training dataset}\par \textbf{dependence} & $L^\text{field}$, $L_\text{noisy}^\text{contrast}$ &	$L^\text{current}$\\
		\hline
	\end{tabular}
\end{table}

\begin{table}[!ht]
	\centering
	\caption{Summary of imaging performances of concerned inverse-scattering-problem solvers}
	\label{tab:summary}
	\begin{tabular}{m{2.5cm}|m{2.3cm}|m{2.4cm}|m{3.7cm}|m{3.5cm}}
		\toprule
		& \textbf{Section} / \textbf{Equation} & \textbf{Advantage} & \textbf{Disadvantage} & \textbf{Suitable} \textbf{scenario}\\
		\hline
		Born iterative method & Section \ref{subsec:analogy} & Broad applicability & Biased estimation & Low requirement on accuracy\\[3pt]
		$L_\text{clean}^\text{contrast}$ & Eq.~\eqref{eq:lossContrast} & Less training time & Weak robustness & High SNR\\[3pt]
		$L_\text{noisy}^\text{contrast}$ & Eq.~\eqref{eq:lossContrast} & Less training time & Fail with weak scatterers & High contrast \\[3pt]
		$L^\text{current}$ & Eq.~\eqref{eq:lossCurrent} & High accuracy & High dependency on training set  &\vspace{2pt} High SNR \& scatterers similar with training set\\[3pt]
		$L^\text{field}$ & Eq.~\eqref{eq:lossField} & Strong robustness & Long training time & High contrast \\
		\bottomrule
	\end{tabular}
\end{table}
According to the above studies, the evaluation results of the DNN solvers are summarized in qualitative terms in Table~\ref{tab:evaluation}. The higher time cost for training with $L^\text{current}$ and $L^\text{field}$ is due to the additional computation of data-fitting errors and the corresponding gradients. The imaging performances of the concerned ISP solvers are briefly described in Table~\ref{tab:summary}, where suitable scenarios are suggested according to the pros and cons of each ISP solver. The traditional method BIM has weakness in reconstruction accuracy and fits scenarios having low requirement on accuracy. Otherwise, the DNN solvers are better options. When SNR is sufficiently high, DNN solvers trained with $L_\text{clean}^\text{contrast}$ and $L^\text{current}$ should be used. The $L^\text{current}$ is especially preferred in favor of higher accuracy and lesser artifacts when the detected scatterers are similar with the training set. If SNR is a concern, the DNN solvers with $L_\text{noisy}^\text{contrast}$ and $L^\text{field}$ are better options but maybe unsuitable for scatterers with low contrast.

\section{Conclusions}
\label{sec:Conclusions}

In this paper, two physics-guided loss functions are proposed to improve the noise robustness and the reconstruction accuracy of deep learning approach for reconstructing scatterers from electric far-field measurement. By incorporating a loss term related to scattered fields in domain of interests (DOI), noise effects are suppressed and superior imaging performances are obtained when the signal-to-noise is low (\SI{5}{\dB} in the paper). Feature-enhanced imaging is achieved when the contrast function is constrained by induced currents in DOI. 

Comparisons of the physics-guided DNNs with conventional methods, which include Born iteration method and back propagation, has been performed. It reveals that the physics-guided DNNs improve the accuracy of reconstructed scatterer profiles, but the performance depends on the test sample being similar to the samples in the training dataset. 
However, we note that the new designs of loss functions are not perfect, and may contribute degradations in some imaging performances. For instance, while features of real scatterers are enhanced by introducing the constraint about induced current, the energy of noise artifacts is also increased. Additionally considering the scattered near-field as a constraint improves robustness against noise, but reduces performance for low-contrast scatterers. More efforts to develop loss functions which can balance the imaging performances and incorporate physical model of scattering in a balanced manner in DNNs are of the present authors' interests.

\bibliographystyle{chicago}
\bibliography{DNN}
\end{document}